\documentclass{article} 
\usepackage{main,times}

\usepackage{amsmath,amsfonts,bm}

\def\eqref#1{equation~\ref{#1}}

\def\1{\bm{1}}

\DeclareMathAlphabet{\mathsfit}{\encodingdefault}{\sfdefault}{m}{sl}
\SetMathAlphabet{\mathsfit}{bold}{\encodingdefault}{\sfdefault}{bx}{n}

\usepackage{hyperref}
\usepackage{url}

\usepackage{algpseudocode}
\usepackage{wrapfig}
\usepackage{booktabs}
\usepackage{caption}
\usepackage{amsmath}
\usepackage{algorithm}
\usepackage{algpseudocode}
\usepackage{amsfonts}
\usepackage{xcolor}
\usepackage{graphicx}

\title{PixelShuffler: A Simple Image Translation Through Pixel Rearrangement}

\author{Omar Zamzam \\
Ming Hsieh Department of Electrical and Computer Engineering\\
University of Southern California\\
Los Angeles, CA 90089, USA \\
\texttt{zamzam@usc.edu} \\
}

\iclrfinalcopy 
\begin{document}

\maketitle

\begin{abstract}

Image-to-image translation is a topic in computer vision that has a vast range of use cases ranging from medical image translation, such as converting MRI scans to CT scans or to other MRI contrasts, to image colorization, super-resolution, domain adaptation, and generating photorealistic images from sketches or semantic maps. Image style transfer is also a widely researched application of image-to-image translation, where the goal is to synthesize an image that combines the content of one image with the style of another. Existing state-of-the-art methods often rely on complex neural networks, including diffusion models and language models, to achieve high-quality style transfer, but these methods can be computationally expensive and intricate to implement. In this paper, we propose a novel pixel shuffle method that addresses the image-to-image translation problem generally with a specific demonstrative application in style transfer. The proposed method approaches style transfer by shuffling the pixels of the style image such that the mutual information between the shuffled image and the content image is maximized. This approach inherently preserves the colors of the style image while ensuring that the structural details of the content image are retained in the stylized output. We demonstrate that this simple and straightforward method produces results that are comparable to state-of-the-art techniques, as measured by the Learned Perceptual Image Patch Similarity (LPIPS) loss for content preservation and the Fréchet Inception Distance (FID) score for style similarity. Our experiments validate that the proposed pixel shuffle method achieves competitive performance with significantly reduced complexity, offering a promising alternative for efficient image style transfer, as well as a promise in usability of the method in general image-to-image translation tasks. \footnote{The implementation code can be found at: \url{https://github.com/OmarSZamzam/PixelShuffler/}.}

\end{abstract}

\begin{figure}[ht]
    \centering
    \includegraphics[width=0.9\textwidth]{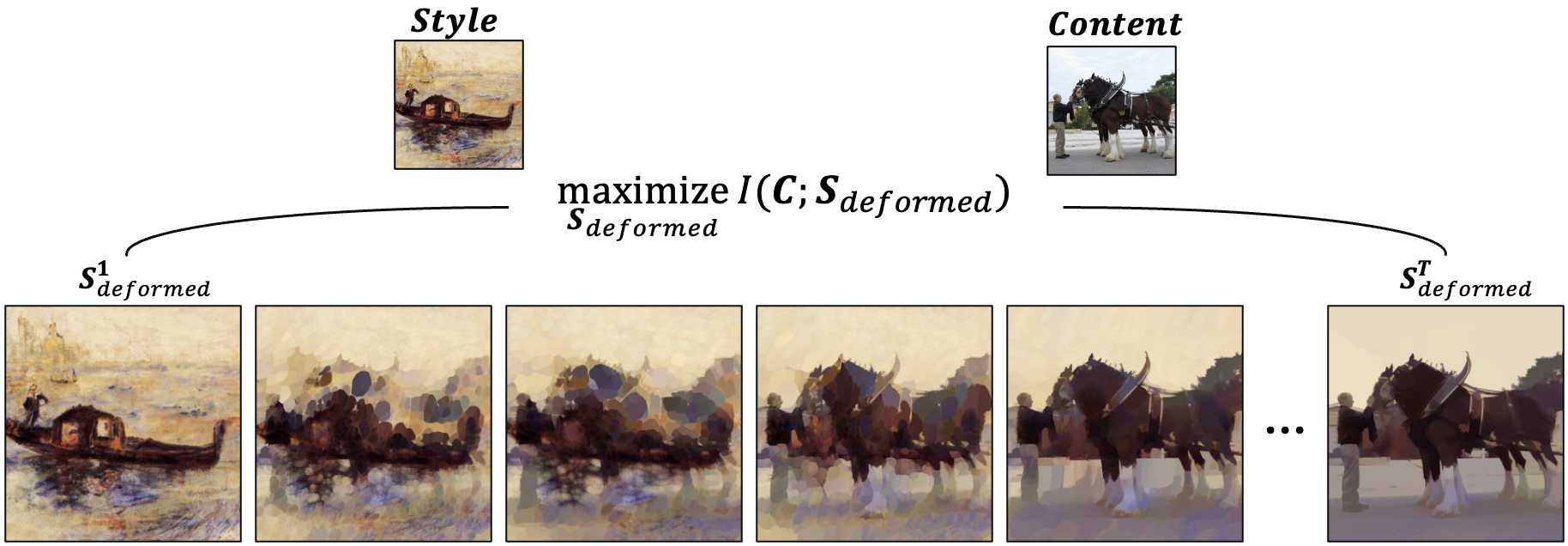}
    \caption{Illustration of how pixel shuffle of the style image guided by mutual information between shuffled image and content image can yield simple and nice style transfer results.}  
\end{figure}

\section{Introduction}

Image-to-image translation is a broad and versatile task in computer vision, encompassing a wide range of applications. These applications span medical image translation (e.g., converting MRI scans to CT scans or other MRI contrasts), image colorization, super-resolution, domain adaptation, and the generation of photorealistic images from sketches or semantic maps. At its core, image-to-image translation aims to map an input image from a source domain to a corresponding image in a target domain, preserving key information from the source domain (such as structural details) while incorporating attributes from the target domain (such as visual characteristics or textures). The goal is to maintain a balance between faithfully retaining essential elements of the source image and applying the desired transformations from the target domain.

One widely studied application of image-to-image translation is style transfer, where the aim is to combine the content of one image with the artistic style of another. This problem has gained substantial attention due to its potential applications in art generation, image editing, and creative industries. The seminal work of \citet{gatys2015neural} laid the foundation for neural style transfer by leveraging convolutional neural networks (CNNs) to separate and recombine content and style features from images. Since then, numerous methods have been developed, ranging from generative adversarial networks (GANs) to more recent diffusion models and language-guided approaches that produce highly stylized outputs.

However, despite the impressive results of these advanced methods, they often come with significant drawbacks, including high computational costs, complex architectures, and the need for extensive training data. Moreover, many of these approaches struggle to preserve the fine structural details of the content image while faithfully transferring the style, often resulting in artifacts or loss of content integrity. As a result, there is a growing demand for simpler and more efficient style transfer techniques that can achieve comparable results without the overhead of complex models.

In this paper, we propose a novel and straightforward image-to-image translation framework, with a specific demonstration in style transfer. Our approach introduces a novel pixel shuffle algorithm that maximizes the mutual information between the shuffled style image and the content image. This method directly manipulates the pixels of the style image, inherently preserving its color palette while ensuring that the structural content of the content image is matched. Our main motivation arises from the observation that shuffling the pixels of a style image can effectively preserve its stylistic characteristics while allowing flexibility to match any structural content. This balance of style and content offers a simpler yet powerful alternative to complex neural architectures. While we demonstrate the effectiveness of this technique in the context of style transfer, the core algorithm is applicable to a wide array of image-to-image translation tasks.

The key contributions of this paper are as follows:
\begin{itemize}
    \item We propose a novel pixel shuffle algorithm for image-to-image translation, demonstrated in style transfer.
    \item We introduce the use of mutual information maximization to balance content preservation and style adaptation, addressing the common issue of content distortion observed in existing methods.
    \item We demonstrate competitive performance with state-of-the-art techniques in style transfer, offering a promising alternative with significantly reduced complexity and broader applicability to other image-to-image translation tasks.
\end{itemize}

\section{Related Work}

Image style transfer has been extensively explored in computer vision, evolving from early neural methods to advanced deep learning techniques. We categorize the related work into five main areas: neural style transfer, generative adversarial networks (GANs), diffusion models, transformer-based approaches, and information theory-based methods.

\textbf{Neural Style Transfer.} The pioneering work by \citet{gatys2015neural} introduced the concept of neural style transfer using convolutional neural networks (CNNs) to separate and recombine content and style features through optimization. This method achieved impressive artistic effects by matching the Gram matrices of feature maps, sparking widespread interest. Subsequent approaches, such as those by \citet{johnson2016perceptual} and \citet{ulyanov2016texture}, improved efficiency by employing feed-forward networks to approximate the optimization process, significantly reducing computational costs. Despite these advancements, these methods often struggle to balance style and content, particularly in preserving fine structural details.

\textbf{Generative Adversarial Networks (GANs).} GANs have been widely adopted in style transfer to enhance realism and stylization quality. Works like \citet{zhu2017unpaired} and \citet{huang2017arbitrary} introduced CycleGAN and Adaptive Instance Normalization (AdaIN), respectively, bringing significant improvements in style transfer through adversarial training and feature modulation. However, GAN-based methods typically require extensive training data and complex architectures, making them computationally expensive and prone to artifacts, particularly in high-resolution scenarios.

\textbf{Diffusion Models and Training-Free Approaches.} Diffusion models have recently gained prominence in image style transfer for their ability to produce high-quality images through iterative refinement. \citet{chung2024style} proposed a training-free style injection method within diffusion frameworks, demonstrating that large-scale diffusion models can adaptively transfer styles without extensive retraining. Their work highlighted the effectiveness of modifying diffusion model priors for controlled stylization, achieving state-of-the-art results with minimal training overhead. Similarly, \citet{chen2024artadapter} developed ArtAdapter, a multi-level style encoder and explicit blending mechanism within a diffusion model, achieving impressive text-to-image style transfer. Despite their flexibility and quality, diffusion-based models remain computationally intensive and complex to implement.

\textbf{Transformer-Based Approaches.} Transformers have emerged as powerful tools in style transfer due to their capacity to capture long-range dependencies. \citet{deng2022stytr2} proposed StyTr2, a transformer-based approach that captures global style information more effectively than CNN-based methods. Their model uses transformers to directly learn the mapping between content and style domains, resulting in enhanced style consistency and better preservation of content details. Meanwhile, recent advancements such as \citet{zhang2024artbank} explore combining transformers with fine-tuning techniques to improve adaptation and personalization in style transfer. These methods represent a growing trend towards leveraging the unique strengths of transformers for style transfer tasks.

\textbf{Statistical Approaches.} Statistical methods offer a novel perspective on style transfer, focusing on content preservation and style adaptation through statistical and information-theoretic principles. \citet{li2017universal} explored aligning content and style distributions using Maximum Mean Discrepancy (MMD) and other statistical measures to achieve style transfer. However, the application of statistical methods and information theory in style transfer remains limited, particularly with the emergence of current diffusion and transformer-based models.

\textbf{Our Approach.} Unlike the aforementioned techniques, our method simplifies style transfer by directly manipulating the style image through a pixel shuffle algorithm – That requires no pre-training. By maximizing mutual information between the shuffled style image and the content image, our approach preserves stylistic elements while maintaining structural details. This direct manipulation bypasses the need for training complex data-hungry neural networks, making it computationally efficient and easy to implement, as demonstrated through competitive performance against state-of-the-art methods.

\section{Methodology}

\subsection{Problem Setup}

Given two input images, a content image \( \mathbf{C} \in \mathbb{R}^{H_c \times W_c \times 3} \) and a style image \( \mathbf{S} \in \mathbb{R}^{H_s \times W_s \times 3} \), the objective of style transfer is to generate an output image \( \mathbf{O} \) that preserves the structural details of the content image \( \mathbf{C} \) while adopting the stylistic characteristics of the style image \( \mathbf{S} \). Traditional approaches often use complex architectures and training strategies to achieve this goal; however, in our method, we simplify the process through a novel approach involving a neural network and mutual information optimization.

\subsection{Proposed Approach}

We introduce a neural network \( \mathcal{N}_{\theta} \) parameterized by \( \theta \) that takes the content image \( \mathbf{C} \) and the style image \( \mathbf{S} \) as inputs and outputs a deformation field \( \mathbf{D} \). This deformation field \( \mathbf{D} \in \mathbb{R}^{H_s \times W_s \times 2} \) represents a pixel-wise transformation that is applied to the style image \( \mathbf{S} \). Mathematically, the deformation field can be expressed as:

\[
\mathbf{D} = \mathcal{N}_{\theta}(\mathbf{C}, \mathbf{S}).
\]

The deformed style image \( \mathbf{S}_{\text{deformed}} \) is then obtained by applying the deformation field \( \mathbf{D} \) to the style image \( \mathbf{S} \). Let \( \mathbf{S}_{\text{deformed}}(x, y) \) be the pixel at coordinates \( (x, y) \) in the deformed style image, then:

\[
\mathbf{S}_{\text{deformed}}(x, y) = \mathbf{S}(x + \mathbf{D}(x, y)_x, y + \mathbf{D}(x, y)_y),
\]

where \( \mathbf{D}(x, y)_x \) and \( \mathbf{D}(x, y)_y \) are the displacement values in the horizontal and vertical directions, respectively.

\subsection{Mutual Information Maximization}

The key idea behind our approach is to maximize the mutual information between the deformed style image \( \mathbf{S}_{\text{deformed}} \) and the content image \( \mathbf{C} \). Mutual information \( I(\mathbf{C}; \mathbf{S}_{\text{deformed}}) \) measures the amount of shared information between the two images, and by maximizing it, we aim to preserve the structure of the content image while embedding the style of \( \mathbf{S} \).

The objective function to be minimized is:

\[
\mathcal{L}_{\text{MI}} = -I(\mathbf{C}; \mathbf{S}_{\text{deformed}}),
\]

where \( I(\mathbf{C}; \mathbf{S}_{\text{deformed}}) \) is the mutual information between the content and the deformed style images. The network \( \mathcal{N}_{\theta} \) is trained iteratively by updating the parameters \( \theta \) to minimize \( \mathcal{L}_{\text{MI}} \). This optimization process ensures that the deformed style image becomes structurally similar to the content image while maintaining the stylistic properties of the original style image.

\subsection{Optimization Process}

The optimization is performed using a gradient-based approach, where the deformation field \( \mathbf{D} \) is iteratively refined. The parameters \( \theta \) of the network \( \mathcal{N}_{\theta} \) are updated according to the gradient of the loss function:

\[
\theta \leftarrow \theta + \eta \nabla_{\theta} \mathcal{L}_{\text{MI}},
\]

where \( \eta \) is the learning rate. The process continues until the loss function converges, producing a deformed style image \( \mathbf{S}_{\text{deformed}} \) that successfully blends the style and content according to the set objective.

\subsection{Final Output}

\begin{figure}[ht]
    \centering
    \includegraphics[width=0.8\textwidth]{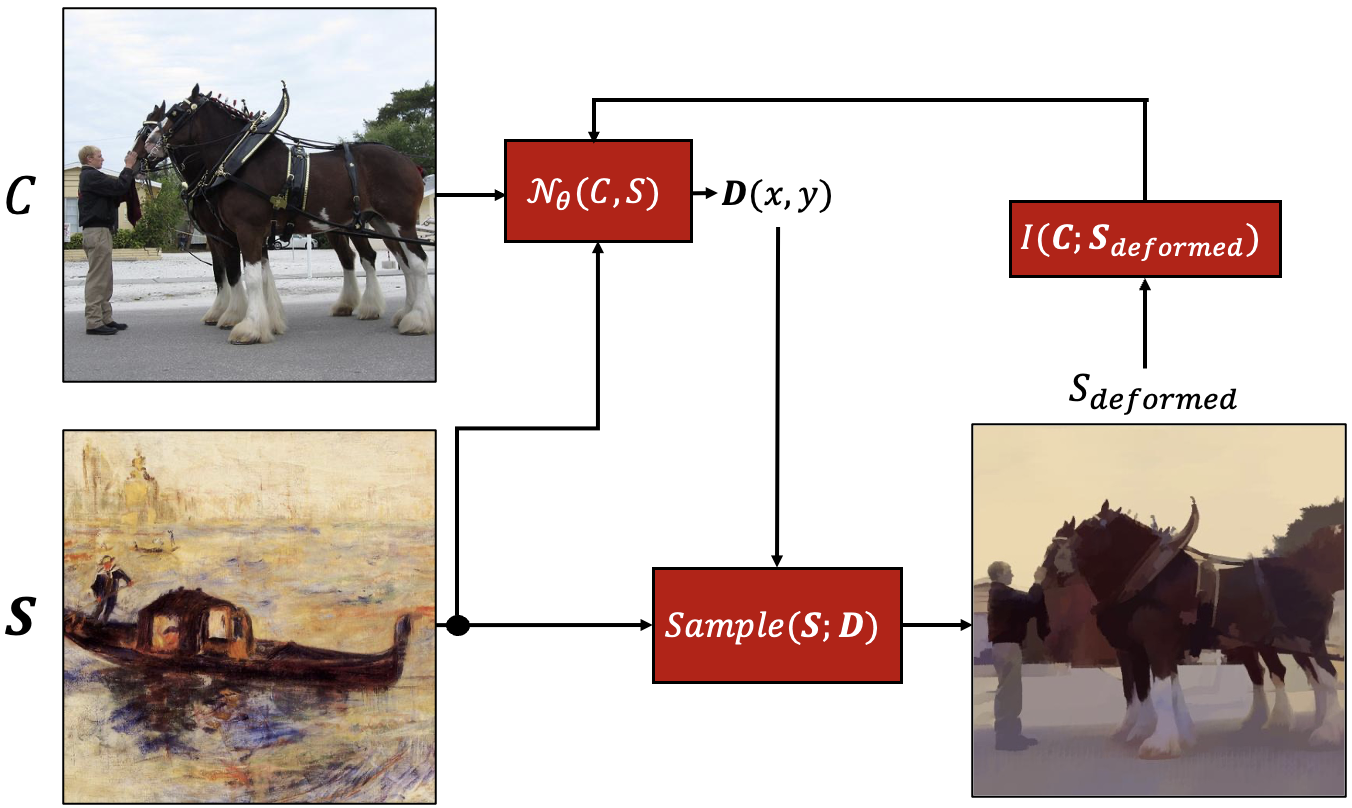}
    \caption{Overview of the proposed style transfer methodology.}  
    \label{method}
\end{figure}

The final output image \( \mathbf{O} \) is the deformed style image \( \mathbf{S}_{\text{deformed}} \) after convergence, which contains the artistic style of \( \mathbf{S} \) while preserving the structural content of \( \mathbf{C} \). This simple yet effective approach bypasses complex deep learning pipelines, achieving competitive style transfer results with reduced computational overhead. Figure \ref{method} shows an overview of the proposed methodology.

\subsection{Ablation Studies}

In addition to the proposed mutual information maximization approach, we experimented with alternative training strategies aimed at directly optimizing for commonly used loss functions in style transfer. Specifically, we tested a variant of our model that directly minimizes the LPIPS loss function for content preservation and a style loss function based on perceptual loss, which has been widely used in previous neural style transfer methods to maintain the style of the style image. 

It is important to note that the proposed training method allows for flexibility in defining custom loss functions that the user aims to optimize. In our experiments, we observed that directly minimizing these loss functions did result in improved quantitative performance across standard evaluation metrics, such as LPIPS for content preservation and FID for style fidelity.

While the combination of LPIPS and perceptual loss yielded competitive results, it did not outperform all state-of-the-art methods. This indicates that direct optimization of such loss functions can sometimes overlook finer interactions between content and style, which are better captured through more nuanced approaches. That said, we believe there is potential for improved results through careful fine-tuning of the weights assigned to each loss function. Finding the right balance between content preservation and style fidelity is key to achieving superior performance, and future work could focus on refining these weights to enhance the results further.

\section{Experiments and Results}

\subsection{Experimental Setup}

To validate the effectiveness of our proposed method, we conducted experiments using a combination of randomly selected natural images from the web and images from the COCO dataset. The initial phase involved fine-tuning the model to select appropriate hyperparameters, including the learning rate and optimizer configuration. We concluded that the Adam optimizer with a learning rate of \(3 \times 10^{-3}\) provided the best balance between convergence speed and stability during training. The architecture of the neural network \(\mathcal{N}_{\theta}\) was a U-Net, which was selected for its efficiency and effectiveness in pixel-wise transformations. The network takes as input the concatenation of the style image \(\mathbf{S}\) and content image \(\mathbf{C}\) and outputs the deformation field \(\mathbf{D}\).

\subsection{Testing Protocol}

For a rigorous evaluation, we followed the testing protocol described in \citet{chung2024style}. Specifically, we used the same 40 style images from the WikiArt dataset and 20 content images from the COCO dataset as referenced in their study. This selection ensures consistency and comparability of our results against other state-of-the-art methods, providing a standardized benchmark for style transfer performance.

\subsection{Evaluation Metrics}

The evaluation of our method focuses on two primary aspects: content preservation and style transfer quality. To measure these, we adopted the metrics used in prior work:

\begin{itemize}
    \item \textbf{Content Preservation:} The Learned Perceptual Image Patch Similarity (LPIPS) metric \cite{zhang2018unreasonable} was employed to evaluate how well the output image \(\mathbf{O}\) retains the structural details of the content image \(\mathbf{C}\). Lower LPIPS values indicate better content preservation.
    \item \textbf{Style Transfer Quality:} The Fréchet Inception Distance (FID) score \cite{martin2017gans} was used to assess the stylistic similarity between the output images \(\mathbf{O}\) and the style images \(\mathbf{S}\). Lower FID scores suggest a higher degree of style fidelity in the output images.
\end{itemize}

These metrics provide a comprehensive evaluation framework that captures both qualitative and quantitative aspects of the style transfer performance.

\subsection{Results}

The results of our experiments are summarized in Table~\ref{tab:comparison} – Built upon the work of \citet{chung2024style}. Our proposed method demonstrates competitive performance against existing state-of-the-art approaches. The LPIPS score confirms that our method effectively preserves the content of the input image, while the FID score indicates that the stylization quality is on par with other advanced techniques, although still having a room for improvement to exceed state-of-the-art methods.

\begin{table}[h]
\centering
\resizebox{\columnwidth}{!}{%
\begin{tabular}{l c c c c c c c c c c c c c c c c}
\toprule
\textbf{Metric} & \textbf{Ours–MI+LPIPS} & \textbf{Ours–MI+PL} & \textbf{Ours–MI} & \textbf{Style Injection} & \textbf{AesPA-Net} & \textbf{CAST} & \textbf{StyTR$^2$} & \textbf{EFDM} & \textbf{MAST} & \textbf{AdaAttn} & \textbf{ArtFlow} & \textbf{AdaConv} & \textbf{AdaIN} & \textbf{DiffuseIT} & \textbf{InST} & \textbf{DiffStyle} \\ 
\midrule
FID $\downarrow$ & 31.81 & 25.45 & 29.359 & \textbf{18.131} & 19.760 & 20.395 & 18.890 & 20.062 & 18.199 & 18.658 & 21.252 & 19.022 & 18.242 & 23.065 & 21.571 & 20.903 \\ 
LPIPS $\downarrow$ & \textbf{0.4271} & 0.7436 & 0.4689 & 0.5055 & 0.5135 & 0.6212 & 0.5445 & 0.6430 & 0.6293 & 0.5439 & 0.5562 & 0.5910 & 0.6076 & 0.6921 & 0.8002 & 0.8931 \\
\bottomrule
\end{tabular}%
}
\caption{Comparison of style transfer performance across different methods using various metrics. The best results are highlighted in bold.}
\label{tab:comparison}
\end{table}

\begin{figure}[ht]
    \centering
    \includegraphics[width=\textwidth]{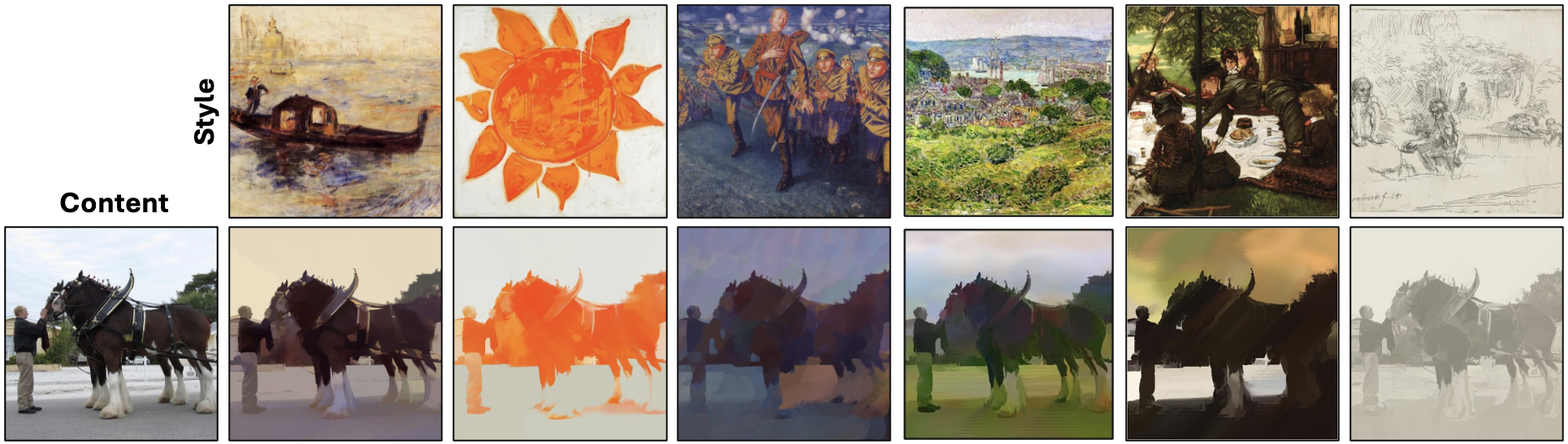}
    \caption{First row are style images in which the content image on bottom left is stylized.}  
    \label{Sample_Images}
\end{figure}

As seen in the table, our method achieves the lowest LPIPS score indicating strong content preservation capabilities, confirming the effectiveness of our mutual information maximization strategy. The FID scores are shown to be outperformed by other methods, which we suppose is due to the lack of explicit style preservation loss. Overall, it is shown that the performance of the proposed method is comparable to all state-of-the-art methods, while maintaining a simple and direct algorithm and implementation that requires minimal resources and data-driven training. Figure \ref{Sample_Images} shows a sample content image and the result of stylizing it using the proposed method into multiple different styles.

\begin{figure}[ht]
    \centering
    \includegraphics[width=\textwidth]{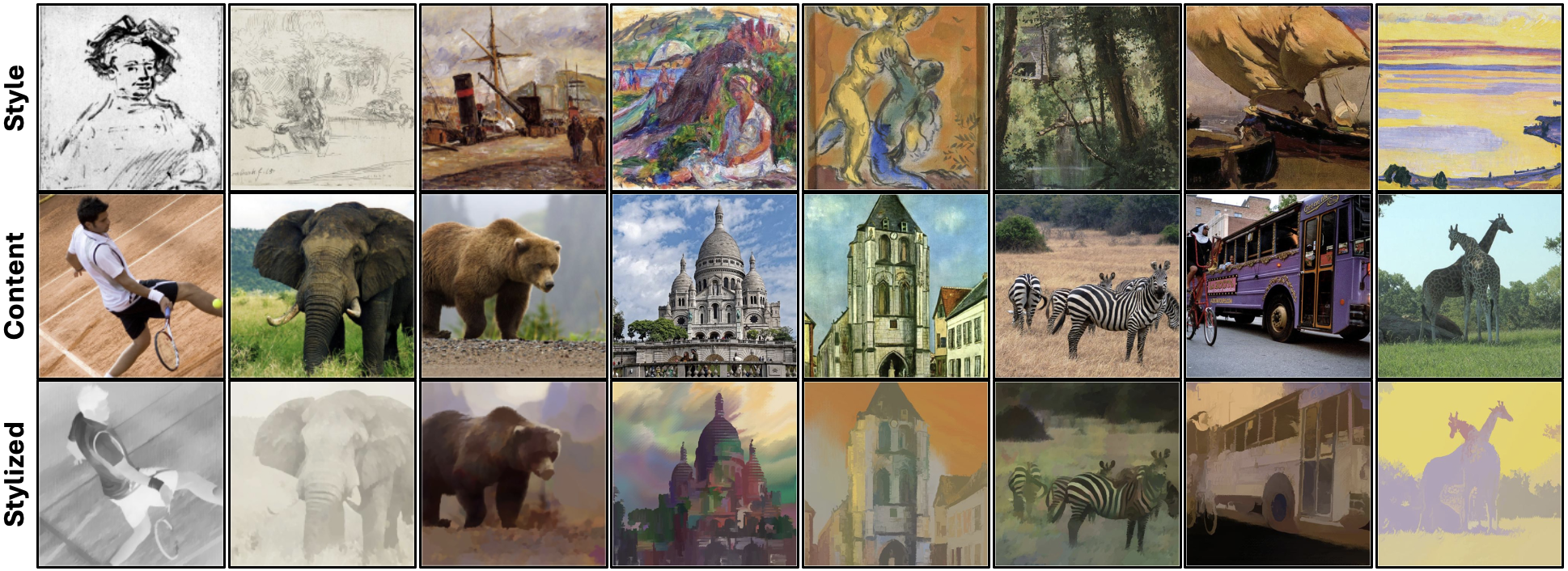}
    \caption{Some example style, content, and corresponding stylized images.}  
    \label{Sample_Images2}
\end{figure}

\subsection{Discussion}

Our results indicate that the proposed PixelShuffler method is a very promising technique for style transfer such that it can target a custom loss function to minimize and outperform other methods in achieving the customized specific goal. The results shown in Table \ref{tab:comparison} show that maximizing mutual information allowed the proposed method to significantly outperform the state-of-the-art methods in maintaining the information from the content image. 

By leveraging a simplified neural architecture and direct pixel manipulation, our approach reduces computational overhead while focusing the performance to improve based on the minimized loss function, suggesting that simpler, more interpretable methods can achieve results comparable to complex deep learning models. Additionally, it is important to note that the proposed method does not require pre-training, which makes it more accessible and readily-applicable in many applications.

These findings open new avenues for exploring alternative style transfer techniques that prioritize efficiency and simplicity, making advanced style transfer accessible without extensive computational resources.

\section{Conclusion}

In this paper, we present a novel and general approach to image-to-image translation, with style transfer serving as a demonstrative application. Our method simplifies the translation problem by directly manipulating the style image through a pixel shuffle algorithm guided by mutual information maximization. This approach, unlike traditional complex neural networks, does not require any pre-training or extensive training or complicated neural architectures, yet achieves competitive results.

While our experiments focused on style transfer, the proposed pixel shuffle method is flexible and can be adapted to various other image-to-image translation tasks, such as medical image translation. The inherent simplicity of our approach makes it accessible and computationally efficient, providing a viable alternative to existing state-of-the-art techniques that often rely on heavy neural architectures.

The success of our method demonstrates that effective image-to-image translation can be achieved through minimalist frameworks, opening up new avenues for research in resource-efficient, interpretable, and flexible image translation techniques. Future work will explore further applications of this method in other image translation domains and refine the model to enhance its performance across different tasks.

\bibliography{main}
\bibliographystyle{main}

\end{document}